\documentclass[conference]{IEEEtran}
\pdfoutput=1 

\usepackage[pdftex]{graphicx}

\usepackage{subfig}

\begin{document}
%
\title{Chittron: An Automatic Bangla Image Captioning System}


\author{ 
\IEEEauthorblockN{ Motiur Rahman$^{1}$,
Nabeel Mohammed$^{2}$, 
Nafees Mansoor$^{3}$ and
Sifat Momen$^{4}$}
\IEEEauthorblockA{$^{1,3}$Department of Computer Science and Engineering, University of Liberal Arts Bangladesh (ULAB), Bangladesh}
\IEEEauthorblockA{$^{2,4}$ Department of Electrical and Computer Engineering, North South University (NSU), Bangladesh}
Email:{nabeel.mohammed@northsouth.edu$^{2}$, nafees@nafees.info$^{3}$, sifat.momen@northsouth.edu$^{4}$}
}


%


\maketitle

\begin{abstract}
Automatic image caption generation aims to produce an accurate description of an image in natural language automatically. However, Bangla, the fifth most widely spoken language in the world, is lagging considerably in the research and development of such domain. Besides, while there are many established data sets to related to image annotation in English, no such resource exists for Bangla yet. Hence, this paper outlines the development of ``Chittron'', an automatic image captioning system in Bangla. Moreover, to address the data set availability issue, a collection of $16,000$ Bangladeshi contextual images has been accumulated and manually annotated in Bangla. This data set is then used to train a model which integrates a pre-trained VGG16 image embedding model with stacked LSTM layers. The model is trained to predict the caption when the input is an image, one word at a time. The results show that the model has successfully been able to learn a working language model and to generate captions of images quite accurately in many cases. The results are evaluated mainly qualitatively. However, BLEU scores are also reported. It is expected that a better result can be obtained with a bigger and more varied data set.
\end{abstract}


%
\IEEEpeerreviewmaketitle

\section{Introduction}
While an image may very well be worth ``a thousand words'', it is hardly ever practical to describe an image with so many. Instead, what is useful in many applications is an adequate description of an image comprising the essential information. Therefore, the automatic methods of image captioning aim to do just that, and have already had major impacts in various fields, \emph{e.g. image search}. Furthermore, it has the potential to influence positive changes in many different areas, including software for disabled individuals, surveillance \& security, human-computer interaction etc.

\begin{figure}[!ht]
     \subfloat[Caption generated: A woman standing in front of a mirror holding a teddy bear.]{
       \includegraphics[width=0.45\textwidth]{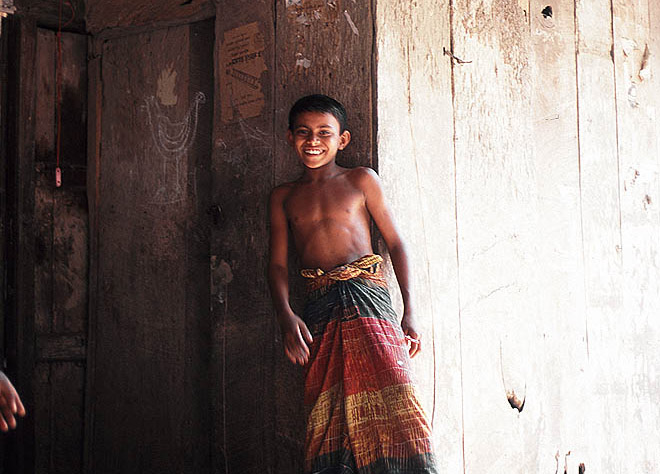}
       \label{westernProblem-a}
     }
     \hfill
     \subfloat[Caption generated: A young boy wearing a bow tie.]{%
       \includegraphics[width=0.45\textwidth]{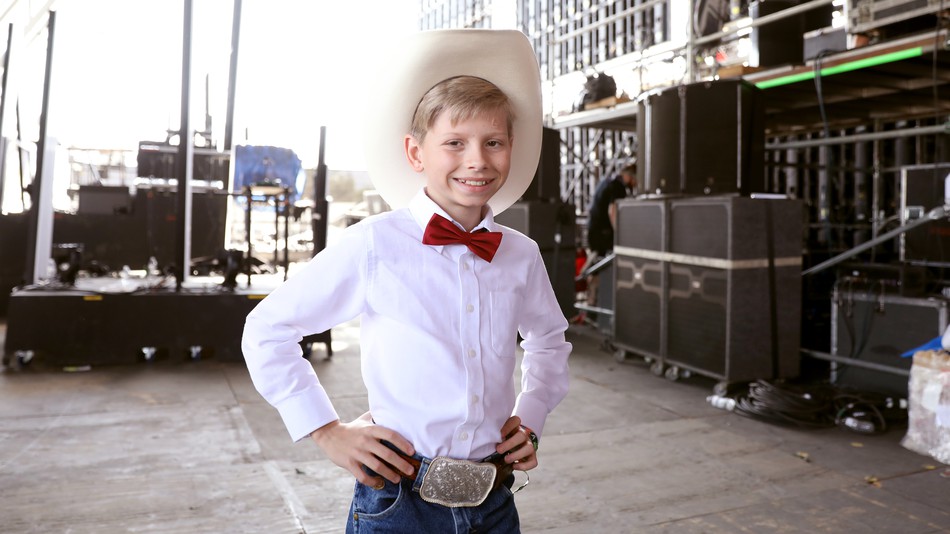}
       \label{westernProblem-b}
     }
     \caption{Example of western bias in existing data sets and captioning models}
     \label{westernProblem}
   \end{figure}

The stark reality is that most of the works in image captioning have concentrated almost exclusively on the English language \cite{vinyals2017show, chen2015mind,rennie2017self}. Additionally, the relevant data-sets, e.g. the MSCOCO \cite{lin2014microsoft},  have a prominent western preference which leads to a two-pronged problem. Firstly, the language in which captions are generated are in English only, and secondly, the data set is not representative of the cultural peculiarities of non-western countries. These very problems exist for generating image captions in Bangla, particularly for images which have a decidedly Bangla geocultural flavor. A simple example of this can be seen in Figure \ref{westernProblem}, where a web service is used to generate captions. The service uses the \emph{im2txt} model trained on the MSCOCO data set and quite clearly the model fails to recognize the image in Figure \ref{westernProblem-a} as a boy wearing a \emph{lungi}, a very common male garb in Bangladesh. In fact, it incorrectly identifies the subject as a female since the attire is identified as women gown. On the other hand, the model shows quite an impressive performance by very precisely describing the image in figure \ref{westernProblem-b}. As depicted in the figure, the model not only rightly identifies the subject to be a \emph{boy}, it also accurately describes what the subject is wearing ( a bow tie ).

While there has been a lot of recent interest in using machine learning on Bangla isolated characters \cite{sharif2016hybrid,7860340,sharif2017comparison,7890867}, there has been no significant work on generating Bangla image captions. Taking into account the present state and the challenges, this paper reports the development of  ``Chittron'', an automatic image annotating system in Bangla. As an initial effort to encounter the unavailability of a proper Bangla geo-contextual image dataset, the data set of $16,000$ images has been produced. It is worthwhile to mention that the images are collected from the public domain of the web with relatively diversified subject matters. Next, a native Bangla speaker annotated each image with a single overly-descriptive Bangla caption. This data set is then used to train a model similar to the \emph{im2txt} model \cite{vinyals2017show}. The proposed model integrates a pre-trained VGG16 image embedding model with stacked LSTM layers. The model is trained to predict the caption when the input is an image, one word at a time. The results show that the model iss successfully able to learn a working language model and to generate captions of images quite accurately in many cases. The results are evaluated mainly qualitatively. However, BLEU scores are also measured and the limitation of the BLEU scores are also discussed. The shortcomings of the current work is also discussed, most of which can be addressed by creating a larger more varied data set.

The rest of the paper is organized as follows. In section II, relevant work in image captioning is discussed. The prepared data set is described later in section III. The model and training details is presented in section IV. In section V, results of the proposed system are presented and discussed. Later, conclusion and future works have been highlighted in section VI.

\section{Relevant work}
This section of the paper presents various existing approaches for image annotation along with data set used in those systems. The work in this area typically takes two different approaches - the traditional approach being predominantly the search or rule based approach where the input image is searched over an image database to find relevant images/words. The problem with this approach is that they tend to be heavily hand-engineered and therefore results tend to be more brittle when facing examples which was not taken into account during the engineering process. The second approach makes heavy use of machine learning techniques, more recently deep learning techniques to learn  the captioning task entirely from data without any human engineered features, except for the network structure itself. Meanwhile, the development of such a system has resulted in a need for image annotation dataset. 

A naive approach towards generating the caption of an image is perhaps to predict words from the image regions and link them up. One of the first image annotation system has been developed back in 1999  \cite{mori1999image}. Later, in 2002 the image captioning task has been re-cast as that of machine translation \cite{duygulu2002object}. However, it turns out that the technique fails to perform properly for a number of reasons which includes simple mapping of an image to a word completely overlooks any kind of relationships that exist between the objects in the image. Moreover, arranging the annotations in the caption becomes very difficult in this system. 

Sentences are richer than a set of individual words since a sentence can describe actions going on in an image. It also shows the relationships among different entities in the image as well as the properties of the objects in the image. Additionally, sentences are also expected to be grammatically correct and natural sounding in the target language. The system presented in \cite{gupta2008beyond} shows that using the spatial relationship between objects improves both the quality of generated annotation and its placings. 
 
Search-based approaches have also seen prominence in the aforementioned area. The main challenge with this approach lies in ranking descriptions for a given image \cite{11}, \cite{8}, \cite{24}. The technique presented in \cite{6} uses object detection techniques to deduce a triplet of scene elements that are then used to generate texts. A similar approach but in conjunction with detections phrases is presented in \cite{19}. This technique detects the objects and their relationships and creates the final description about the image. Another approach presented in \cite{16} uses a combination of template-based and a more complex detection technique to generate the texts. Language parsing models have also been used \cite{23, 1, 17, 18, 5} to generate texts. Ranking descriptions for a given image have been one of the challenges of this approach. One suggested solution towards this is co-embedding the images and text in the same vector space\cite{13, 29} to allow easy queries. Against a particular image query, descriptions have sought that lie closest to the image in the embedding space. 

The work presented in \cite{vinyals2017show} pursued the second machine learning approach where a pretrained convolutional neural netowrk\cite{12} is used to extract a rich image embedding vector, which is then used in a recurrent neural network (RNN) ( particularly stacked LSTM layers )\cite{10}, for sequence modeling. This model treats the captions as sequences to be predicted one word at a time and aims to learn a languages model directly from the data. This model is inspired by the successes of sequence generation witnessed in neural machine translation \cite{3, 2, 30}. Other similar works include that of \cite{chen2015mind} which uses a recurrent visual representation of an image and \cite{rennie2017self} which uses deep reinforcement learning for the image captioning task. These systems are trained end-to-end, as in, no human intervention or human engineering is done except the network architecture. These networks are then trained to generation captions directly from images as inputs, using different strategies. 

All the mentioned approaches are developed for English languages. Hence, this research identifies the necessity of a Bangla Captioning System. In this paper, the model trained and used to generate Bangla image captions is similar to that of \cite{vinyals2017show}.

\section{BanglaLekha-ImageCaptions: The data set}

\begin{figure}[!ht]
     \subfloat[In English: A man walking with a basket on his head on a road with trees on both sides]{%
       \fbox{\includegraphics[width=0.45\textwidth]{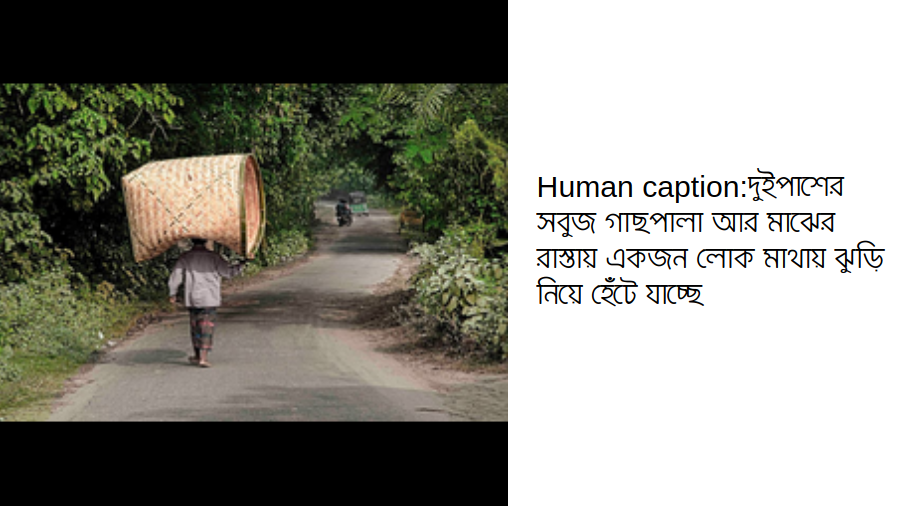}}
       \label{example-b}
     }
     \hfill
     \subfloat[In English: Two baby boys and a baby girl are on a boat picking water lilies]{%
       \fbox{\includegraphics[width=0.45\textwidth]{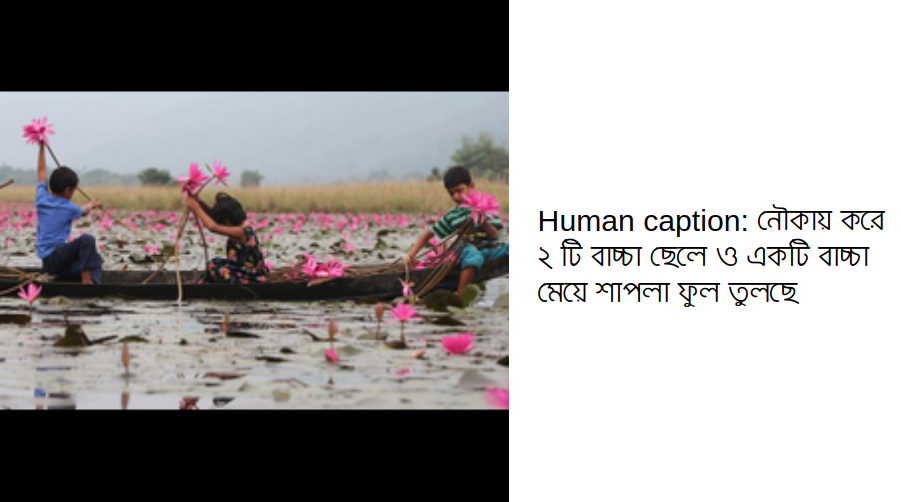}}
       \label{example-c}
     }
     \hfill
     \caption{Some examples of BanglaLekha-ImageCaptions. Bangla captions shown with the images, translated English captions shown as the labels.}
     \label{caption-examples}
\end{figure}

This is the second data set collected in the BanglaLekha series. The first is BanglaLekha-Isolated \cite{biswas2017banglalekha}, which concentrated on images of isolated Bangla characters. This offering is quite different in its nature, but no less important. Data sets like MSCOCO has $200,000$ images, with $5$ captions per image. BanglaLekha-ImageCaptions is significantly smaller in size, with only $16,000$ images, all collected from the public domain of the web with relatively diversified subject matters. Almost all the images are tied to Bangladesh in some way, with some being relevant to the wider Indian Subcontinental context. For each image, a native Bangla speaker is tasked with writing a caption. As multiple captions per image were not possible due to different constraints, the annotator was instructed to write overly-descriptive captions. Figure \ref{caption-examples} shows a few example images and their captions in Bangla. The English translation of the captions is also shown in the labels of each sub-figure.

Moreover, analyses of the captions reveal that the data set has 6035 unique Bangla words, where words with the same root but different prefixes and/or suffixes are counted as different words. Numerals are also counted as individual words in this scheme. 

\section{Model and Training Details}

This section of the paper discusses the model that has been used in the proposed system. The discussion has been split into two parts where one subsection concentrates on the model preparations and the other one focuses on the training environment. 

\subsection{Model Description}
In the proposed model the image captioning task can be broadly divided into two parts, (a) extracting relevant image features, and (b) generating a language description using the features. The proposed model is very similar to existing successful models, e.g. \emph{im2txt, NeuralTalk} \cite{vinyals2017show}, as in it uses a pre-trained model to extract image embeddings. Later, the model uses a one-word-at-a-time strategy to predict the caption from stacked LSTM layers. The widely used VGG16 \cite{simonyan2014very} model which is pre-trained on the Imagenet dataset, is used as the pre-trained image model in the proposed system with some slight adjustments. However, other more recent models (e.g. resnet, resnext ) might give better results, but the available hardware did not facilitate training with them.

\begin{figure*}[!ht]

\begin{center}
     \includegraphics[scale=0.5]{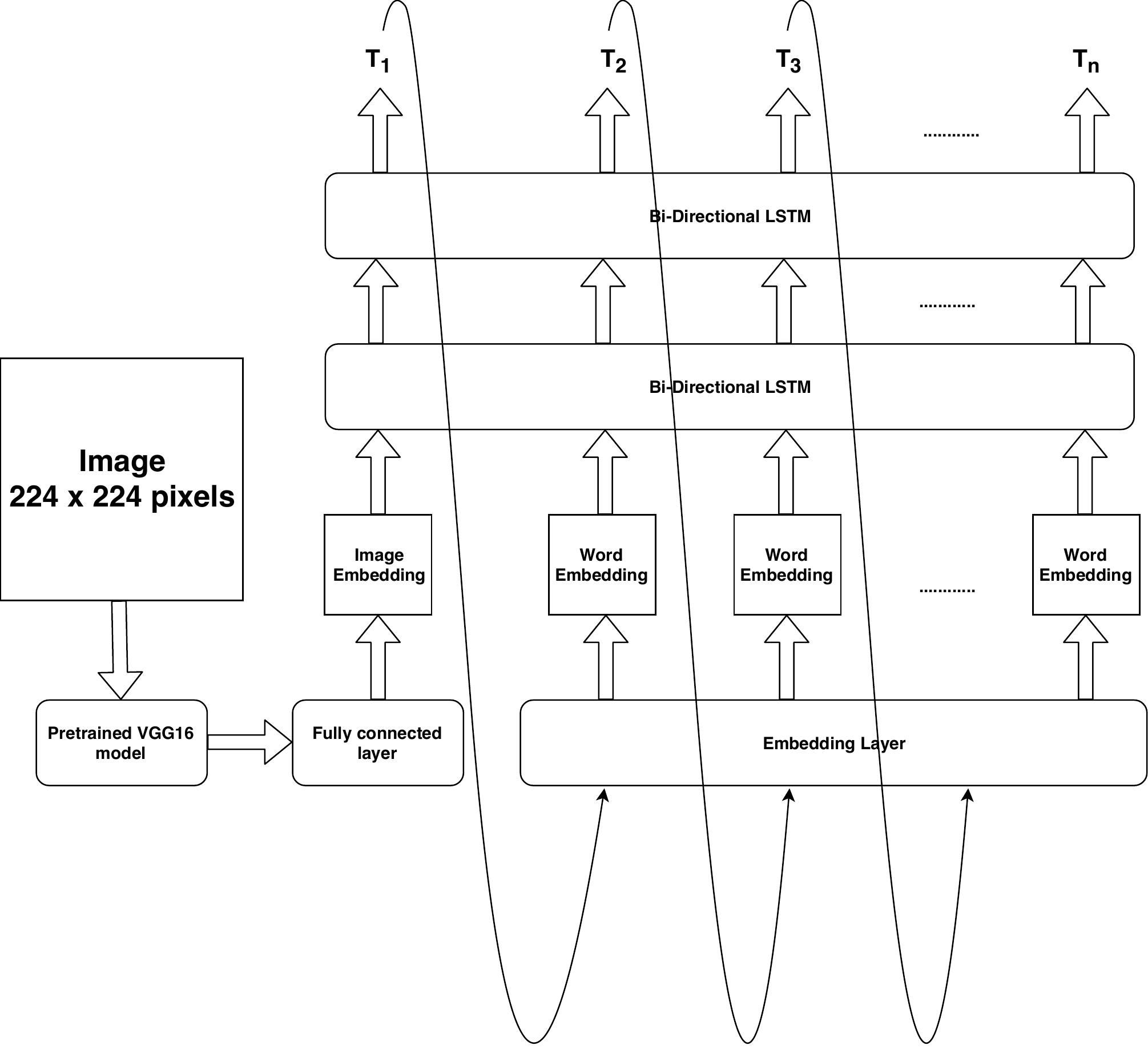}
     \caption{The image captioning model employed}
     \label{model-diagram}
\end{center}

\end{figure*}

Figure \ref{model-diagram} depicts the details of the model used in the training phase. The last layer of the pre-trained VGG16 model is discarded, as the requirement is to get an image descriptor, not probability distributions. The model has two inputs: the first input is the image itself, which is fed into the VGG16 model, and the second input is a sequence of tokens (corresponding to each unique word in the vocabulary). An embedding layer accepts the tokens as input and generates corresponding word embeddings \cite{mikolov2013distributed}. The output of the second last layer of the VGG16 model is reduced to 512 dimensions through a fully connected layer. Next, the model concatenates with the output of the embedding layer, whose output is also 512 dimensions. The concatenated data forms the sequence data input for the stacked LSTM layers. In total, the sequence data fed into the stacked LSTM layers is a sequence of $n+1$ embeddings, where $n$ is the maximum length of the generated caption. This model uses bidirectional LSTMs for both the LSTM layers.

The model is set up such that it takes an image as input and a sequence of tokens, and it is meant to predict the next token in the caption. The work in this paper predicts captions up to 10 tokens in length. Therefore, for each image-caption pair, the caption is first truncated or extended to 10 tokens. The truncation has consequences which are further discussed in the later section of the paper. The extension is easily done by padding the tokens with the \emph{unknown} token. Once, all the captions are of the required length, for each image-caption pair 10 training data points are created. The first data point has no tokens, i.e. all tokens are the \emph{unknown} token, in which case the model is expected to predict the first word from the image embedding alone. In each data point an increasing number of tokens are added, so that the last data point has the first $n-1$ tokens of the caption, expecting the model to predict the last $n^{th}$ token. This is further explained in Table \ref{ModelInputTable}, which shows how the input samples would be arranged if the maximum sequence length is set to five. Here $I_i$ is the embedding vector of image $i$, derived from the output of the VGG16 model. $U$ is the "unknown'' token. $T_{ik}$ is the $k^{th}$ token in the caption of image $i$. For this setup to work, the image embedding must have the same dimension as the word embeddings calculated from each token using the Embedding layer.

\begin{table}[ht]
\caption{Example of training sequence data} 
\label{ModelInputTable}
\centering 
\begin{tabular}{ | c | c | c | c | c |}
  \hline
  $I_{i}$ & $U$ & $U$ & $U$ & $U$ \\
  \hline
  $I_{i}$ & $T{i1}$ & $U$ & $U$ & $U$ \\
  \hline
  $I_{i}$ & $T{i1}$ & $T{i2}$ & $U$ & $U$ \\
  \hline
  $I_{i}$ & $T{i1}$ & $T{i2}$ & $T{i3}$ & $U$ \\
  \hline
  $I_{i}$ & $T{i1}$ & $T{i2}$ & $T{i3}$ &$T{i4}$ \\
  \hline
\end{tabular}

\end{table}

The model is trained on $15,700$ images from the collected data set, resulting in $15,700 \times 10$ training samples, following the scheme outlined above. $300$ images are considered as the test data. Ideally, more test images should be used, however, given the small size of the entire data set, coupled with the fact that it has only a single caption per image. However, it is required to train the model with a larger dataset comprising numerous images and captions to enable the model to learn a working language model.

The proposed model is trained end-to-end using the back propagation method. Basic Stochastic Gradient Descent is used to minimize the categorical cross entropy of the output of the stacked LSTM layers.

\section{Results and discussion}

Both quantitative and qualitative results are discussed in this section. Quantitative results are presented in terms of the BLEU score \cite{papineni2002bleu}, which is a widely used metric for evaluating machine translation systems and has been applied in image captioning work as well. The reasons why the BLEU scores give misleading results are discussed with the qualitative assessment of results. As the output of the model is captions in Bangla, English translations are supplied along with an indication of the correctness level of the caption language, from a grammatical perspective.

300 images were used to test the trained caption model. The model achieved an average BLEU score of 2.5, which is admittedly extremely poor. However, BLEU scores are usually calculated in cases where multiple reference sentences are available. As the current data set only has a single caption per image, BLEU scores are not necessarily a good indication of performance. The qualitative assessment presented below will further bolster this notion. 

\begin{figure}[!ht]
     \subfloat[]{
       \fbox{\includegraphics[width=0.45\textwidth]{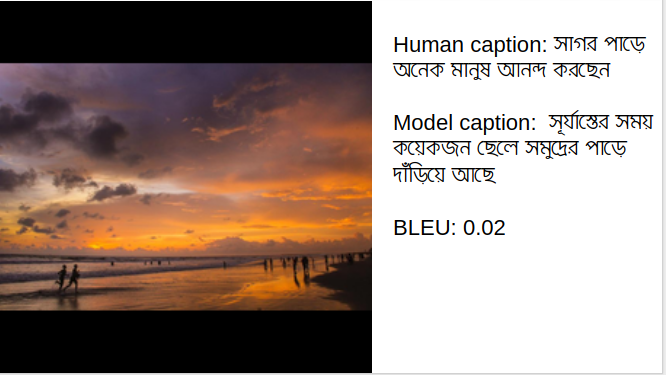}}
       \label{cg-a}
     }
     \hfill
     \subfloat[]{%
       \fbox{\includegraphics[width=0.45\textwidth]{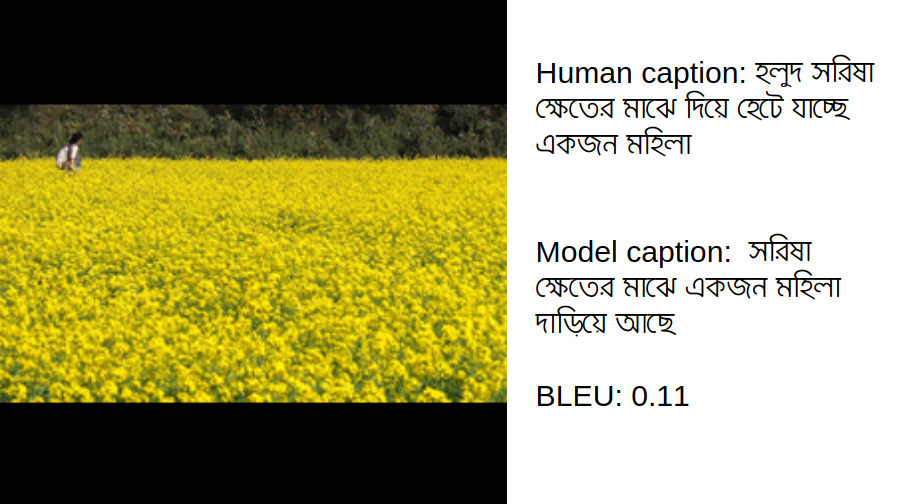}}
       \label{cg-b}
     }
     \hfill
     \subfloat[]{%
       \fbox{\includegraphics[width=0.45\textwidth]{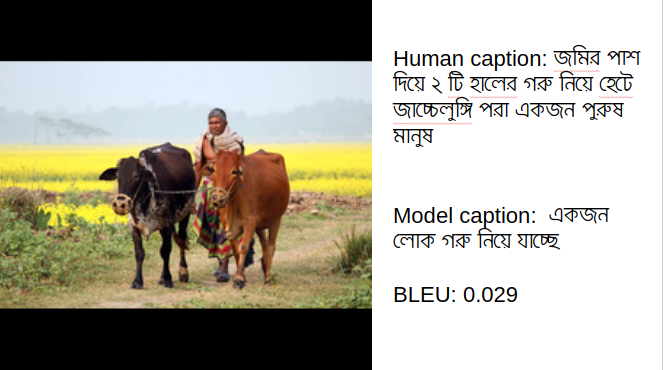}}
       \label{cg-c}
     }
     \caption{Examples of appropriate generated captions}
     \label{caption-examples-good}
\end{figure}

\begin{figure}[!ht]
     \subfloat[]{
       \fbox{\includegraphics[width=0.45\textwidth]{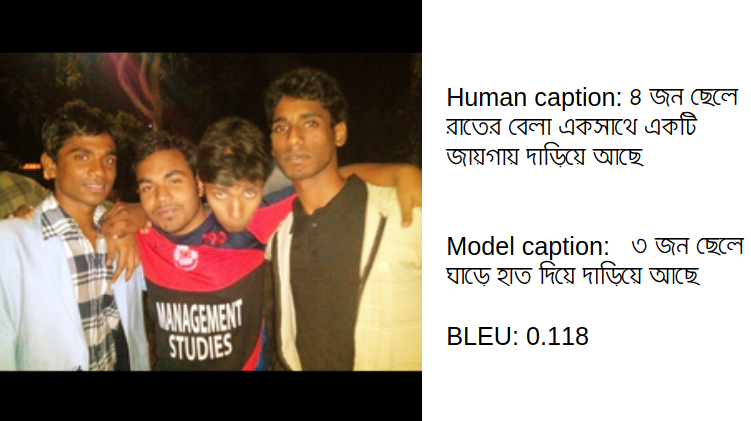}}
       \label{cb-a}
     }
     \hfill
     \subfloat[]{%
       \fbox{\includegraphics[width=0.45\textwidth]{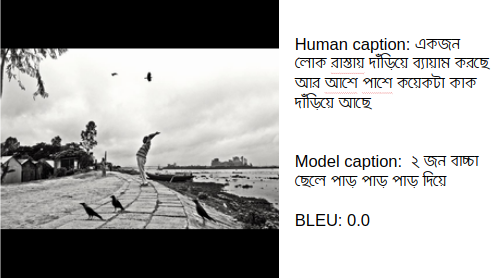}}
       \label{cb-b}
     }
     \hfill
     \subfloat[]{%
       \fbox{\includegraphics[width=0.45\textwidth]{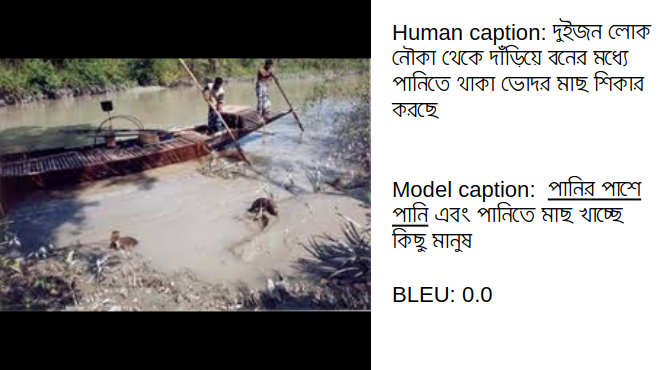}}
       \label{cb-c}
     }
     \caption{Examples of inappropriate or wrong generated captions}
     \label{caption-examples-bad}
\end{figure}

Figure \ref{caption-examples-good} and \ref{caption-examples-bad} shows three images each with their corresponding human annotated caption as well as their model generated captions. In all cases, the BLEU score is unsatisfactory. However, for the three images in Figure \ref{caption-examples-good}, even though the BLEU scores are low, the generated captions are grammatically correct and appropriate. For those who cannot read Bangla, translations are given in Table \ref{caption-examples-good-table}.

\begin{table}[ht]
\caption{Translations of captions in Figure \ref{caption-examples-good}} 
\label{caption-examples-good-table}
\centering 
\begin{tabular}{ | c | p{3cm} | p{3cm} |}
  \hline
  Subfigure & Human annotation & Model Generated \\
  \hline
  \hline
  \ref{cg-a} & A lot of people are having fun in the sea shore & A few boys are standing on the sea shore at sunset\\
  \hline
  \ref{cg-b} & A woman is walking through a field of yellow mustard & A woman is standing in he middle of a mustard field\\
  \hline
  \ref{cg-c} & A man wearing a lungi is walking by some farmland with two cows & A man is taking some cows\\
  \hline
  
\end{tabular}

\end{table}

While the BLEU scores are quite low, the model predictions are indistinguishable from what may be generated by a human. Figure \ref{caption-examples-bad} shows three images where the generated captions either have mistakes or are entirely inappropriate. These also have low BLEU scores, with the lowest being zero. Table \ref{caption-examples-bad-table} shows the corresponding translations and includes a comment about what is wrong with the generated caption.

\begin{table}[ht]
\caption{Translations of captions in Figure \ref{caption-examples-bad}} 
\label{caption-examples-bad-table}
\centering 
\begin{tabular}{ | c | p{1.8cm} | p{1.8cm} | p{1.8cm}|}
  \hline
  Subfigure & Human annotation & Model Generated & Comment\\
  \hline
  \hline
  \ref{cb-a} & Four boys are standing together in a place at night. & Three boys are standing together with hands on each other's shoulders. & Wrong count of boys\\
  \hline
  \ref{cb-b} & A man is exercising on the street and a few crows are standing nearby. & Two baby boys on the shore. & The word for shore repeated three times, Incomplete sentence, Not relevant to the image\\
  \hline
  \ref{cb-b} & Two men are standing on a boat and fishing for Bhodor fish in the water & Water next to the water and some people are eating fish in the water & Nonsensical start of the caption, Wrong description\\
  \hline
\end{tabular}

\end{table}

These examples demonstrate that it is possible to learn a working language model of Bangla entirely from human annotated captions. As is the case in previous English-based studies, the model seems to learn constructs of the target languages grammar entirely from the caption data presented. This can be easily attributed to the stacked LSTM layers as shown in \cite{hassan2016learning}. The shortcomings presented in Figure \ref{caption-examples-bad} can be addressed by adding more captions per image in the data set and also increasing the number of images so that a larger domain of objects and actions are included for the model to learn. The case of the incomplete sentence can be easily traced back to the fact that training was only done using the first 10 tokens of the captions and larger captions were truncated. This has lead to a situation where the model learns to create incomplete sentences. The data set also has the weakness that the majority of the images have subject looking at the camera, so the model has not been exposed to cases like Figure \ref{cb-c} very often, leading to misdirection.

\section{Conclusion}
This paper reports on the development of ``Chittron'', a system to automatically generating Bangla image captions using Deep Neural Networks and the collection of  BanglaLekha-ImageCaptions data set consists of 16000 images, with a single overly-descriptive captions per image. This data set is used to train a model employing a pre-trained VGG16 model and stacked LSTM layers. The VGG16 model was used to extract a rich description of the image content as an image embedding vector. This is then used in a model with stacked LSTM layers, which in turn was trained to predict the captions one word at a time. As LSTM layers are effective for sequential data, they are employed to learn a working Bangla language model so that the generated captions can appear to be in natural language.

Both quantitative and qualitative evaluation are done to analyze the results. BLEU scores used for quantitative evaluation is found to be not appropriate for this particular case as the data set only includes a single reference caption per image. Qualitative evaluation demonstrates that in cases the generated captions are on par with human annotations. Such cases clearly demonstrate the capacity of such models to not only learn an effective language model but one which can be conditioned upon image content, making it effective for image captioning.  However, as the data set is small, there are also cases where the model makes grammatical mistakes or generates entirely inappropriate captions. These can be remedied by curating a larger, more varied data set with multiple captions per image. Which will also allow for more effective quantitative assessment of the work. This is scheduled to be done in the future.

\bibliographystyle{IEEEtran}
\bibliography{bibliographies/IEEEexample}
%

\end{document}